\documentclass{article}

\usepackage{microtype}
\usepackage{graphicx}
\usepackage{subcaption}
\usepackage{booktabs} 
\usepackage{enumitem}

\usepackage{hyperref}

\usepackage[preprint]{icml2026}

\usepackage{amsmath}
\usepackage{amssymb}
\usepackage{mathtools}
\usepackage{amsthm}
\usepackage{amsfonts}
\usepackage{makecell}
\usepackage{multirow}
\usepackage{xcolor}
\usepackage{colortbl}
\usepackage{bbding}

\definecolor{lightgray}{gray}{0.92}

\usepackage[capitalize,noabbrev]{cleveref}

\theoremstyle{plain}

\theoremstyle{definition}

\theoremstyle{remark}

\usepackage[textsize=tiny]{todonotes}

\icmltitlerunning{Look Inward to Explore Outward: Learning Temperature Policy from LLM Internal States via Hierarchical RL}
\begin{document}

\twocolumn[
  \icmltitle{Look Inward to Explore Outward: Learning Temperature Policy from LLM Internal States via Hierarchical RL}



  \icmlsetsymbol{equal}{*}
  \icmlsetsymbol{projectleader}{\textdagger}
  \icmlsetsymbol{corresp}{\Envelope}

  \begin{icmlauthorlist}
    \icmlauthor{Yixiao Zhou}{zju,sii}
    \icmlauthor{Yang Li}{projectleader,yli}
    \icmlauthor{Dongzhou Cheng}{sii,seu}
    \icmlauthor{Hehe Fan}{corresp,zju}
    \icmlauthor{Yu Cheng}{corresp,sii,cuhk}
  \end{icmlauthorlist}

  \icmlaffiliation{zju}{Zhejiang University}
  \icmlaffiliation{seu}{Southeast University}
  \icmlaffiliation{yli}{Independent Researcher, yangli951102@gmail.com}
  \icmlaffiliation{cuhk}{The Chinese University of Hong Kong}
  \icmlaffiliation{sii}{Shanghai Innovation Institute}

  \icmlcorrespondingauthor{Hehe Fan}{hehefan@zju.edu.cn}
  \icmlcorrespondingauthor{Yu Cheng}{chengyu@cse.cuhk.edu.hk}

  \icmlkeywords{Machine Learning, ICML}

  \vskip 0.3in
]



\printAffiliationsAndNotice{\icmlProjectLeader\icmlCorresp}

\begin{abstract}
Reinforcement Learning from Verifiable Rewards (RLVR) trains large language models (LLMs) from sampled trajectories, making decoding strategy a core component of learning rather than a purely inference-time choice. Sampling temperature directly controls the exploration--exploitation trade-off by modulating policy entropy, yet existing methods rely on static values or heuristic adaptations that are decoupled from task-level rewards. We propose \emph{Introspective LLM}, a hierarchical reinforcement learning framework that learns to control sampling temperature during generation. At each decoding step, the model selects a temperature based on its hidden state and samples the next token from the resulting distribution. Temperature and token policies are jointly optimized from downstream rewards using a coordinate ascent scheme. Experiments on mathematical reasoning benchmarks show that learned temperature policies outperform fixed and heuristic baselines, while exhibiting interpretable exploration behaviors aligned with reasoning uncertainty.
\end{abstract}

\section{Introduction}
Reinforcement Learning from Verifiable Rewards (RLVR) has significantly advanced LLM reasoning by optimizing models directly against task-level rewards, yielding substantial gains across domains such as mathematical reasoning~\cite{shao2024deepseekmath,guo2025deepseek}, code generation~\cite{wang2024enhancing}, and agentic tasks~\cite{team2025kimi,zhang2025landscape}. Unlike supervised fine-tuning, where sampling primarily affects inference diversity, RLVR relies on the diversity of generated trajectories to drive policy improvement and sample efficiency. This dependency transforms the decoding strategy, particularly the sampling temperature $\tau$, into a critical training-time component rather than a mere hyperparameter. By scaling the logits of the token distribution, temperature directly modulates policy entropy to govern the essential trade-off between exploring high-variance rollouts and exploiting stable generations.


Despite the central role of temperature in RLVR, most existing approaches rely on either \emph{static} temperature schedules or \emph{heuristic} forms of adaptation. A single temperature is typically selected and applied uniformly across prompts, tokens, and training stages~\cite{shao2024deepseekmath,guo2025deepseek,yu2025dapo}. Recent empirical studies highlight the fragility of this design choice. For example, POLARIS~\cite{an2025polaris} shows that the optimal sampling temperature varies substantially across base models and training phases, and that reinforcement learning progressively reduces model entropy, leading to a contraction of the effective exploration space. Similarly, ~\citet{yang2025let} demonstrates that exploration is not uniformly beneficial across a sequence: high entropy is most valuable at early reasoning steps, while later tokens benefit from lower stochasticity.

More recent methods introduce dynamic adjustments based on hand-crafted signals, such as token entropy~\cite{zhang2024edt,zhuang2025exploring}, KL divergence~\cite{chang2023kl}, or predefined token categories~\cite{zhu2024hot}. While effective in specific regimes, these approaches hard-code assumptions about when and how exploration should occur, limiting their ability to adapt across tasks, prompts, and evolving model capabilities.

Crucially, heuristic temperature rules are \emph{decoupled} from the learning objective: they do not receive direct feedback from task-level rewards and therefore cannot improve through experience. As a result, temperature selection remains a brittle design choice, even though it directly shapes the trajectories used for policy optimization. This raises a natural question: \emph{Can a language model learn to control its own sampling behavior in order to maximize downstream reward?}

We answer this question by enabling the language model to "sense" its own generation context and modulate its exploration strategy accordingly. Specifically, we allow the model to observe its internal state during generation and decide how much stochasticity to inject at each step. This transforms sampling temperature from a rigid external hyperparameter into a context-aware policy that is optimized jointly with token generation under a unified reward signal.

We refer to this framework as \emph{Introspective LLM (IntroLLM)}. Formally, it formulates adaptive temperature selection as a hierarchical reinforcement learning problem, in which temperature decisions and token decisions are learned together under a shared reward signal. At each decoding step, generation proceeds under two coupled policies:
\begin{itemize}[leftmargin=*,topsep=0pt,itemsep=0pt]
    \item a \textbf{temperature policy} $\pi_\phi(\tau_t \mid h_t, \tau_{t-1})$ that selects the sampling temperature based on the model’s internal hidden state; and
    \item a \textbf{token policy} $\pi_\theta(y_t \mid h_t, \tau_t)$ that samples the next token from the temperature-adjusted distribution.
\end{itemize}
This hierarchical structure enables the model to introspect its own uncertainty and generation context, dynamically deciding when to explore through higher-entropy sampling and when to exploit through more deterministic generation. Unlike prior approaches that rely on hand-crafted entropy thresholds or fixed schedules, these decisions are learned directly from data. Crucially, temperature choices are optimized with respect to the same task-level reward signal used to train the language model, allowing exploration strategies to adapt as the model’s capabilities evolve over the course of training and to vary across prompts and token positions.

Jointly learning temperature and token policies introduces nontrivial optimization challenges. To address this, we adopt a coordinate ascent training scheme where both policies are optimized via Group Relative Policy Optimization (GRPO)~\cite{shao2024deepseekmath}. We further model temperature selection as a \emph{mixed discrete--continuous policy}: the policy first decides whether to change the temperature, and only then samples a new value if a change is made. This design allows the model to learn both \emph{when} temperature adaptation is beneficial and \emph{how much} adjustment is required, while avoiding unnecessary variance from frequent changes.

Experiments on mathematical reasoning benchmarks demonstrate that learned temperature policies consistently outperform both fixed-temperature baselines and heuristic adaptive methods. Beyond quantitative gains, analysis of the learned policies reveals interpretable patterns: higher temperatures are allocated to uncertain multi-step reasoning segments, while lower values are reserved for execution-heavy tokens such as numerical computation, factual retrieval, and final answer synthesis. Notably, these behaviors mirror insights from prior empirical studies~\cite{wang2025stabilizing,an2025polaris}, but emerge naturally from reward-driven learning rather than manual design. 

In summary, our main contributions are:
\begin{itemize}[leftmargin=*,topsep=0pt,itemsep=0pt]
    \item We identify the decoding strategy, specifically sampling temperature, as a critical lever for exploration in RLVR and argue that it should be optimized rather than fixed.
    \item We formulate adaptive temperature selection as a hierarchical reinforcement learning problem and propose a principled joint training framework.
    \item We introduce a mixed discrete–continuous temperature policy that learns when and how to adapt stochasticity, balancing exploration flexibility with training stability.
    \item We demonstrate consistent improvements over static and heuristic adaptive temperature methods on reasoning benchmarks, with interpretable learned behaviors.
\end{itemize}

\begin{figure*}[t]
    \centering
    \includegraphics[width=1\linewidth]{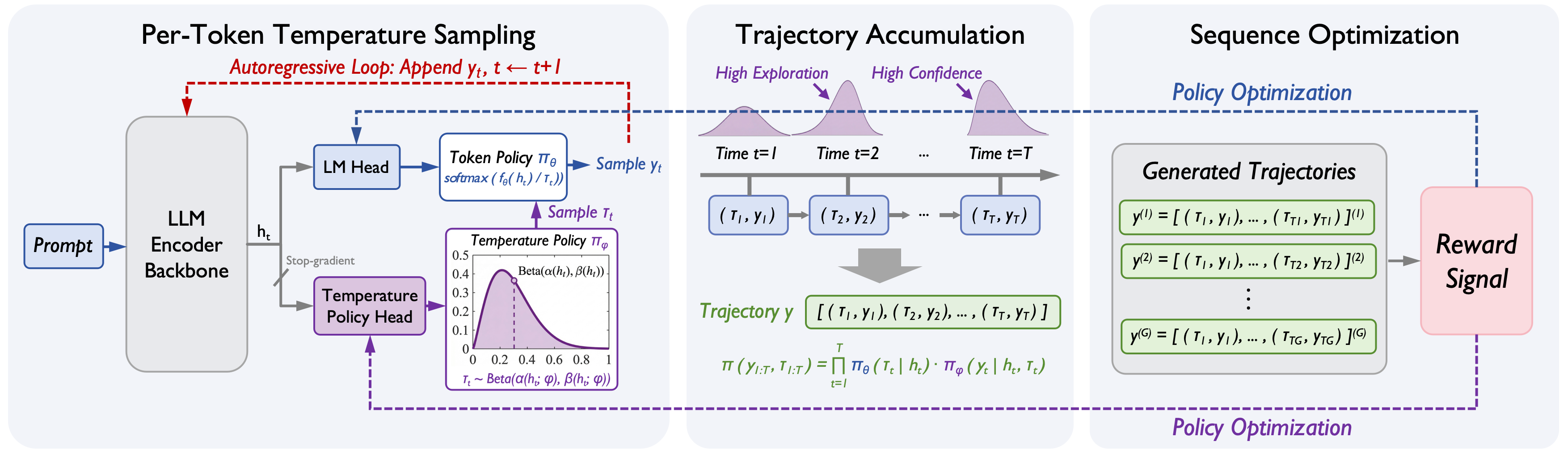}
    \caption{\textbf{Overview of the IntroLLM framework.} At each decoding step, a temperature policy $\pi_\phi$ observes the hidden state $h_t$ and selects a sampling temperature $\tau_t$, which then conditions the token policy $\pi_\theta$ to generate the next token $y_t$. Both policies are jointly optimized via reinforcement learning from verifiable task rewards.}
    \label{fig:arch}
    \vspace{-10pt}
\end{figure*}

\section{Preliminary}

\textbf{Reinforcement Learning with Verifiable Rewards.}
An autoregressive LLM defines a stochastic policy $\pi_\theta$ that samples tokens $y_t \sim \pi_\theta(y_t \mid h_t)$ to generate a sequence $y$ given prompt $x$, aiming to maximize the expected reward $R(x, y)$ over a prompt distribution $\mathcal{D}$:
\begin{equation}
\max_\theta J(\theta) := \mathbb{E}_{x \sim \mathcal{D}, y \sim \pi_\theta(\cdot \mid x)} [R(x, y)].
\end{equation}
To mitigate reward hacking in complex reasoning~\cite{gao2023scaling,wang2025beyond_reward_hacking}, Reinforcement Learning from Verifiable Rewards (RLVR) employs deterministic, rule-based signals for reliable feedback~\cite{guo2025deepseek}. Optimization typically follows the policy gradient:
\begin{equation}
\textstyle \nabla_\theta J(\theta) = \mathbb{E} \left[ \sum_{t=1}^{T} \nabla_\theta \log \pi_\theta(y_t \mid h_t) \, A_t \right],
\end{equation}
where $A_t$ is the advantage and $T$ is the sequence length. To avoid the computational overhead of separate value models required by PPO~\cite{schulman2017proximal}, Group Relative Policy Optimization (GRPO)~\cite{shao2024deepseekmath} samples a group of $G$ trajectories for each prompt and computes sequence-level advantages within the group:
\begin{equation}\label{eq:adv}
\textstyle A_i = \frac{R(x, y^{(i)}) - \mathrm{mean}_{k \in 1:G} R(x, y^{(k)})}{\mathrm{std}_{k \in 1:G} R(x, y^{(k)})}.
\end{equation}
The policy is then updated via a clipped surrogate objective to ensure training stability:
\begin{equation}\label{eq:grpo}
\mathcal{L}(\theta) = \mathbb{E}\Big[\min \big(r_t(\theta) A_t, \;\text{clip}(r_t(\theta), 1-\epsilon, 1+\epsilon) A_t \big) \Big],
\end{equation}
where $r_t(\theta) = \frac{\pi_\theta(y_t \mid h_t)}{\pi_{\theta_{\text{old}}}(y_t \mid h_t)}$ is the importance ratio. This approach enables efficient credit assignment for RLVR without the computational burden of value networks.

\textbf{Sampling Temperature.}
Temperature sampling~\cite{ackley1985learning} is a widely used method to control the stochasticity of token selection. At step $t$, the model produces logits $\ell_t \in \mathbb{R}^{|V|}$ over the vocabulary $V$, which are rescaled by a temperature $\tau > 0$:
\begin{equation}
\textstyle \pi_\theta(y_t = v \mid [x, y_{<t}]; \tau) = \frac{\exp(\ell_{t,v} / \tau)}{\sum_{v' \in V} \exp(\ell_{t,v'}/\tau)},
\end{equation}
where $v \in V$. Higher temperatures ($\tau > 1$) flatten the distribution, encouraging exploration by making less probable tokens more likely. Lower temperatures ($\tau < 1$) sharpen the distribution, favoring exploitation of high-probability tokens. In RLVR, temperature directly affects the diversity of trajectories, and therefore exploration during training. Unlike prior work that fixes or heuristically schedules $\tau$, in our framework the temperature is treated as a learnable control variable, optimized jointly with token generation.

\section{Problem Formulation}\label{sec:problem}

Standard RLVR employs static or heuristic temperature schedules that cannot adapt to task heterogeneity. Empirical studies reveal three key limitations: (i) different reasoning steps require different exploration levels—reasoning pivots benefit from high temperature while factual tokens require low temperature~\cite{cui2025entropy,wang2025stabilizing}; (ii) optimal temperature varies across prompts and difficulty levels; (iii) exploration requirements evolve during training as model capabilities improve~\cite{an2025polaris}.

We address these limitations by formulating temperature selection as a learnable sequential decision process. At each step $t$, a temperature policy $\pi_\phi(\tau_t \mid h_t, \tau_{t-1})$ selects sampling temperature $\tau_t$ based on hidden state $h_t$, while a token policy $\pi_\theta(y_t \mid h_t, \tau_t)$ generates the next token. The joint process factorizes as:
\begin{equation}
\textstyle \pi(y, \tau \mid x) = \prod_{t=1}^{T} \pi_\phi(\tau_t \mid h_t, \tau_{t-1}) \, \pi_\theta(y_t \mid h_t, \tau_t),
\end{equation}
where $y = (y_1, \dots, y_T)$ and $\tau = (\tau_1, \dots, \tau_T)$ are the token and temperature sequences. The objective is to maximize the expected reward:
\begin{equation}
J(\theta, \phi) = \mathbb{E}_{x \sim \mathcal{D}} \, \mathbb{E}_{\tau, y \sim \pi(\cdot \mid x)} \big[ R(x, y) \big].
\end{equation}
This hierarchical formulation enables: (i) reward-driven exploration optimized from task feedback, (ii) context-aware adaptation via hidden state conditioning, and (iii) principled separation of content generation and exploration control. In the following section, we expound on the architectural design and joint optimization procedure for this framework.

\section{Introspective LLM}

As illustrated in \cref{fig:arch}, IntroLLM instantiate the hierarchical formulation through two coupled policies: (i) a \textbf{temperature policy} $\pi_\phi$ that observes hidden state $h_t$ and selects sampling temperature $\tau_t$, and (ii) a \textbf{token policy} $\pi_\theta$ that generates the next token $y_t$ from a $\tau_t$-modulated distribution. These operate sequentially at each decoding step. Training follows a standard RL pipeline: sample trajectories with both temperature and token decisions, evaluate via verifiable rewards, and update both policies jointly using policy gradients. The following subsections detail the token policy conditioning (\S\ref{sec:token}), temperature policy design (\S\ref{sec:temp}), and joint optimization procedure (\S\ref{sec:opt}).

\subsection{Token Policy with Temperature Conditioning}\label{sec:token}

The token policy generates the next token from a distribution modulated by the temperature policy. Specifically, given hidden state $h_t$, the model produces pre-softmax logits $\ell_t \in \mathbb{R}^{|\mathcal{V}|}$, which are rescaled by the selected temperature $\tau_t$ to obtain the token distribution: $\pi_\theta(y_t \mid h_t, \tau_t) = \mathrm{Softmax}\!\left( \frac{\ell_t}{\tau_t} \right)$. This temperature-conditioned formulation enables adaptive entropy modulation, directly addressing the token-level heterogeneity identified in \S\ref{sec:problem}. During training, $\{\tau_t\}$ is treated as fixed environmental parameters, allowing the token policy to focus on content generation while the temperature policy controls exploration intensity. The joint optimization procedure is detailed in \S\ref{sec:opt}.

\subsection{Temperature Policy Design}\label{sec:temp}

The temperature policy $\pi_\phi(\tau_t \mid h_t)$ controls exploration by mapping the model's internal hidden states to dynamic sampling temperatures. This introspective mechanism enables the model to increase stochasticity at exploration steps while maintaining stability during exploitation. To balance expressive control with optimization stability, we formulate $\pi_\phi$ as a mixed discrete–continuous policy. This design allows the model to learn both \emph{when} to adjust temperature and \emph{how much} to adjust it, avoiding unnecessary variance from frequent changes while maintaining adaptation flexibility.

\textbf{Architectural Parameterization.}\quad
We implement the temperature policy as a lightweight MLP head $f_\phi$ branching from the final decoder layer. Given hidden representation $h_t \in \mathbb{R}^d$ at step $t$, the policy head projects it through a low-dimensional bottleneck to produce a 3-dimensional control vector $\mathbf{u}_t = [u_c, u_\alpha, u_\beta]^\top$:
\begin{equation}
\mathbf{u}_t = \mathbf{W}_2 \cdot \text{ReLU}(\mathbf{W}_1 h_t + \mathbf{b}_1) + \mathbf{b}_2,
\end{equation}
where $\mathbf{W}_1 \in \mathbb{R}^{\frac{d}{2} \times d}$ and $\mathbf{W}_2 \in \mathbb{R}^{3 \times \frac{d}{2}}$ are linear projections. This architecture ensures temperature decisions are conditioned on the full semantic and uncertainty context encoded in $h_t$, enabling the policy to dynamically balance exploration and exploitation.

\textbf{Mixed Discrete–Continuous Action Space.}\quad
Temperature selection proceeds through a two-stage stochastic process:

Stage 1: Update Decision.
The policy first decides whether to adjust temperature via a Bernoulli random variable:
$
c_t \sim \text{Bernoulli}(\sigma(u_c)),
$
where $\sigma(\cdot)$ is the sigmoid function. When $c_t=0$, temperature remains constant ($\tau_t = \tau_{t-1}$), reducing optimization variance by avoiding redundant adjustments when the current temperature is appropriate.

Stage 2: Value Sampling.
Conditioned on $c_t=1$, the policy samples a new temperature intensity $z_t$ from a Beta distribution $\mathcal{B}(\alpha, \beta)$, which is naturally bounded in $[0, 1]$ and provides flexible bimodal and unimodal behaviors:
\begin{equation}
z_t \sim \text{Beta}(\text{softplus}(u_\alpha) + \epsilon, \text{softplus}(u_\beta) + \epsilon),
\end{equation}
where $\epsilon = 10^{-6}$ ensures numerical stability. 
The final sampling temperature is obtained via affine transformation to $[\tau_{\min}, \tau_{\max}]$:
$
\tau_t = \tau_{\min} + z_t \cdot (\tau_{\max} - \tau_{\min}).
$

\textbf{Joint Log-Likelihood.}\quad
For policy gradient updates (\S\ref{sec:opt}), we compute the joint log-probability of a realized sample $(c_t, z_t)$ as:
\begin{equation}
\log \pi_\phi(\tau_t \mid h_t) = \log P(c_t) + c_t \log p(z_t \mid \alpha, \beta),
\end{equation}
where $c_t \in \{0, 1\}$ gates the continuous component. This factorization enables gradient computation while maintaining the coupling between discrete and continuous decisions.

\begin{table*}[t]
    \centering
    \caption{\textbf{Performance comparison on reasoning benchmarks.} For AIME24, AMC23, MATH-500, and Average, we report both \textbf{Avg@8} and \textbf{Pass@8}. \textbf{Bold} indicates best performance; \underline{Underline} indicates the second best.}
    \vspace{-5pt}
    \label{tab:combined_results_final}
    
    \small 
    \setlength{\tabcolsep}{3.8pt} 
    
    \begin{tabular}{lccccccccccc}
        \toprule
        \multirow{2}{*}{\textbf{Method}} & \multicolumn{2}{c}{\textbf{AIME24}} & \multicolumn{2}{c}{\textbf{AMC23}} & \multicolumn{2}{c}{\textbf{MATH-500}} & \textbf{Minerva} & \textbf{Olympiad} & \textbf{Omni} & \multicolumn{2}{c}{\textbf{Average}} \\
        \cmidrule(lr){2-3} \cmidrule(lr){4-5} \cmidrule(lr){6-7} \cmidrule(lr){11-12}
        & \textbf{Avg@8} & \textbf{Pass@8} & \textbf{Avg@8} & \textbf{Pass@8} & \textbf{Avg@8} & \textbf{Pass@8} & \textbf{Avg@8} & \textbf{Avg@8} & \textbf{Avg@8} & \textbf{Avg@8} & \textbf{Pass@8} \\
        \midrule
        
        \multicolumn{12}{c}{\textit{\textbf{Qwen3-1.7B-Base}}} \\
        \cmidrule[0.25pt](lr){1-12}
        Base & 1.67 & 10.00 & 18.12 & 65.00 & 31.30 & 74.80 & 7.12 & 12.69 & 8.23 & 13.19 & 49.93 \\
        GRPO ($\tau=1.0$) & 7.08 & 16.67 & \underline{37.50} & 67.50 & 65.25 & 84.00 & 17.28 & 28.78 & 17.40 & 28.88 & 56.06 \\
        GRPO ($\tau=0.6$) & 6.67 & 20.00 & 35.00 & 67.50 & 64.60 & 85.60 & 15.17 & 28.23 & 16.66 & 27.72 & 57.70 \\
        GRPO ($\tau=1.2$) & 7.50 & \underline{23.33} & 35.63 & \underline{70.00} & \textbf{68.35} & 85.40 & \underline{17.33} & \underline{29.93} & 17.43 & 29.36 & 59.58 \\
        EAD (1.2$\rightarrow$0.1) & 7.08 & \underline{23.33} & \underline{37.50} & \underline{70.00} & 67.03 & \underline{85.80} & \underline{17.33} & 29.64 & \underline{17.90} & \underline{29.41} & \underline{59.71} \\
        TAMPO (0.6-1.5) & \underline{7.92} & \underline{23.33} & 35.00 & \underline{70.00} & 65.45 & 84.80 & 16.64 & 27.41 & 17.28 & 28.28 & 59.38 \\
        \textbf{IntroLLM (Ours)} & \textbf{9.58} & \textbf{26.67} & \textbf{39.06} & \textbf{72.50} & \underline{67.53} & \textbf{86.20} & \textbf{18.29} & \textbf{29.95} & \textbf{18.43} & \textbf{30.47} & \textbf{61.79} \\ 
        \cmidrule[0.25pt](lr){1-12}
        
        \multicolumn{12}{c}{\textit{\textbf{Qwen3-4B-Base}}} \\
        \cmidrule[0.25pt](lr){1-12}
        Base & 6.67 & 20.00 & 27.19 & 72.50 & 48.68 & 86.40 & 12.50 & 24.30 & 14.94 & 22.38 & 59.63 \\
        GRPO ($\tau=1.0$) & 14.17 & 33.33 & 50.00 & 80.00 & 79.10 & 92.00 & 21.51 & 42.36 & 23.88 & 38.50 & 68.44 \\
        GRPO ($\tau=0.6$) & 11.67 & 30.00 & 50.94 & \underline{82.50} & 77.60 & 91.40 & 21.09 & 41.06 & 23.41 & 37.63 & 67.97 \\
        GRPO ($\tau=1.2$) & 14.58 & 30.00 & \underline{51.88} & \underline{82.50} & \underline{80.03} & 92.00 & 22.61 & \underline{44.12} & \underline{24.35} & 39.60 & 68.17 \\
        EAD (1.2$\rightarrow$0.1) & 14.17 & \underline{36.67} & \underline{51.88} & 77.50 & 80.00 & \underline{92.20} & \textbf{23.90} & 43.69 & 24.32 & \underline{39.66} & 68.79 \\
        TAMPO (0.6-1.5) & \underline{15.00} & 33.33 & 50.62 & \underline{82.50} & 78.23 & 91.80 & 22.79 & 42.43 & 23.92 & 38.83 & \underline{69.21} \\
        \textbf{IntroLLM (Ours)} & \textbf{19.17} & \textbf{40.00} & \textbf{52.50} & \textbf{87.50} & \textbf{80.73} & \textbf{93.00} & \underline{23.48} & \textbf{44.25} & \textbf{25.34} & \textbf{40.91} & \textbf{73.50} \\
        \bottomrule
    \end{tabular}   
    \vspace{-5pt}
\end{table*}

\subsection{Joint Policy Optimization}\label{sec:opt}
Jointly optimizing both policies presents a challenge: simultaneous updates can lead to non-stationary learning dynamics where each policy's gradient depends on the other's changing behavior. We address this through coordinate ascent, alternating between token policy updates (with fixed temperatures $\tau$) and temperature policy updates (with fixed tokens $y$). This decoupling stabilizes training while maintaining 
policy interaction through the shared reward $R(x, y)$. Algorithm~\ref{alg:introllm_final} details the complete procedure.

\textbf{Hierarchical Rollout.}\quad
During rollout, generation proceeds via a nested stochastic process. At each step $t$, the temperature policy samples $(c_t, z_t)$ to determine $\tau_t$, and the token policy samples $y_t$ conditioned on $\tau_t$. We persist the complete trajectory $\mathcal{T} = \{(c_t, z_t, \tau_t, y_t)\}_{t=1}^T$ in the experience buffer, where 
$\{y_t\}$ enables reward evaluation and $\{(c_t, z_t, \tau_t)\}$ enables on-policy likelihood recomputation.

\textbf{On-Policy Likelihood Recomputation.}\quad
To compute policy gradients, we must evaluate the log-probabilities of the actions taken during rollout under the \emph{current} policy parameters. During each training iteration, we perform a single forward pass to recompute likelihoods for both policies:

1. Token Policy: We compute $\log \pi_\theta(y_t \mid h_t, \tau_t)$ by scaling token logits with the cached temperatures $\{\tau_t\}$ from the stored trajectory. This ensures gradients reflect the exact exploration context of the original rollout.

2. Temperature Policy: We compute $\log \pi_\phi(c_t, z_t \mid h_t)$ for the recorded temperature actions. This allows the exploration strategy to be reinforced based on the sequence-level rewards it produced.

Both likelihoods are then used in the GRPO objective to update the respective policies via coordinate ascent.

\textbf{Coordinate Ascent Optimization.}\quad
We optimize $\theta$ and $\phi$ via coordinate ascent, alternating updates within each GRPO iteration. Given a group of trajectories $\mathcal{T} = \{(c^{(i)}, z^{(i)}, \tau^{(i)}, y^{(i)})\}_{i=1}^G$:

1. Token Policy Update. Fixing $\tau$ as constants, we update $\theta$ to maximize reward conditioned on the realized exploration:
\begin{equation*}
\textstyle \theta \leftarrow \arg\max_\theta \mathbb{E}_{\mathcal{T}} \left[ \frac{1}{G} \sum_{i=1}^G \sum_{t=1}^T \mathcal{L}_{i,t}^{\text{token}}(\theta \mid \tau_{t}^{(i)}) \right],
\end{equation*}
where $\mathcal{L}^{\text{token}}$ is the GRPO clipped objective from Equation~\eqref{eq:grpo} applied to token actions $y_t$, using the importance ratio $r_t^\theta = \pi_\theta(y_t | h_t, \tau_t) / \pi_{\theta_{\text{old}}}(y_t | h_t, \tau_t)$.

2. Temperature Policy Update. Fixing tokens $y$ as constants, we update $\phi$ to optimize the exploration strategy that produced those rewards:
\begin{equation*}
\textstyle \phi \leftarrow \arg\max_\phi \mathbb{E}_{\mathcal{T}} \left[ \frac{1}{G} \sum_{i=1}^G \sum_{t=1}^T \mathcal{L}_{i,t}^{\text{temp}}(\phi \mid y_{t}^{(i)}) \right],
\end{equation*}
where $\mathcal{L}^{\text{temp}}$ applies the same GRPO objective (Equation~\eqref{eq:grpo}) to temperature actions $(c_t, z_t)$ with importance ratio $r_t^\phi = \pi_\phi(c_t, z_t | h_t) / \pi_{\phi_{\text{old}}}(c_t, z_t | h_t)$.

Both updates share the same trajectory-level advantage $A_i$ (Equation~\eqref{eq:adv}), computed from the final reward $R(x, y^{(i)})$ and normalized across the group. This shared advantage ensures both policies receive consistent credit assignment: if a trajectory achieves high reward, both the temperature decisions and token choices that produced it are reinforced proportionally. This alternating scheme enables the token policy to learn optimal content generation under diverse exploration regimes, while the temperature policy learns to identify high-reward exploration strategies. By isolating gradient flows, we mitigate the non-stationarity of simultaneous updates, achieving 
stable convergence.

\section{Experiments}


\textbf{Datasets and Benchmarks.}
We train all models on the MATH~\cite{hendrycks2021measuring} 
training set and evaluate on a comprehensive suite of reasoning tasks: AIME 2024, 
AMC 2023, MATH-500~\cite{lightman2023let}, Minerva Math~\cite{lewkowycz2022solving}, 
OlympiadBench~\cite{he2024olympiadbench}, and Omni-Math~\cite{gao2024omni}.

\textbf{Baselines.}
We compare IntroLLM against three categories of baselines: 
(i) Standard GRPO: GRPO training with fixed temperatures $\tau \in \{0.6, 1.0, 1.2\}$ to establish performance bounds for non-adaptive sampling; 
(ii) Heuristic Adaptive (EAD)~\cite{yang2025let}: A pre-defined annealing schedule that monotonically decays temperature to balance early-stage exploration and late-stage convergence; 
(iii) Sequence-level Adaptive (TAMPO)~\cite{dang2026temperature}: A learnable meta-policy that optimizes a static temperature for the entire sequence generalization.

\textbf{Implementation Details.}
All models are trained for 4 epochs with batch size 128, learning rate $1 \times 10^{-6}$, group size $G=8$, and maximum response length 3072 tokens. For IntroLLM, we initialize the temperature policy head to a maximum-entropy state ($P(c_t=1) = 0.5$, $z_t \sim \text{Uniform}(0, 1)$) with temperature bounds restricted to $[\tau_{\text{min}}, \tau_{\text{max}}] = [0.6, 1.5]$. 

\textbf{Evaluation Metrics.}
We report \textbf{Avg@8} (expected Pass@1 performance across multiple samples) and \textbf{Pass@8} to assess both reasoning accuracy and generation diversity. All methods use their respective temperature strategies with maximum response length 3072. Further details are provided in Appendix~\ref{app:implementation}.

\subsection{Main Results}

\textbf{Mathematical Reasoning Performance.}
\cref{tab:combined_results_final} presents results on mathematical reasoning benchmarks. We first examine the impact of fixed temperature choices. Performance varies substantially across different temperatures for GRPO, with a 1.6-2.0\% gap between best and worst configurations. Moreover, no single temperature is optimal—the best configuration varies across benchmarks and model scales (e.g., $\tau=1.2$ for AIME24 on 4B, $\tau=1.0$ for AMC23 on 1.7B). This demonstrates that temperature significantly affects performance, yet static schedules cannot adapt to varying problem characteristics.

Adaptive baselines show improvements: EAD (pre-defined annealing $1.2\rightarrow0.1$) achieves 29.41\%/39.66\% average Avg@8 (1.7B/4B), while TAMPO (sequence-level learned temperature) reaches 28.28\%/38.83\%. IntroLLM, with learned token-level adaptive control, achieves 30.47\%/40.91\%—outperforming all baselines across both scales. Gains are most pronounced on challenging benchmarks: on AIME24 (4B), IntroLLM reaches 19.17\%, surpassing the best baseline by 4.17\%; on Omni-Math (4B), it achieves 25.34\%, outperforming all methods by ${\sim}$1-2\%. Notably, improvements correlate with difficulty. The 4B model gains 4.17\% on AIME24 but only 0.7\% on MATH-500, suggesting that adaptive temperature particularly benefits complex reasoning where exploration is most critical. Moreover, benefits increase with model scale. Compared to the second-best method, Avg@8 gains grow from 1.06\% to 1.25\%, while Pass@8 gains more than double from 2.08\% to 4.29\%, indicating larger models particularly benefit from learned exploration in discovering diverse solutions.

Beyond improving expected accuracy, IntroLLM significantly enhances solution diversity. It achieves Pass@8 of 61.79\%/73.50\%, surpassing the second-best by 2.08\%/4.29\%. The diversity advantage is especially prominent on AMC23 (4B), where IntroLLM reaches 87.50\%—a 5.0\% gain over the strongest baseline. Across all benchmarks, IntroLLM simultaneously achieves the highest Avg@8 and Pass@8, demonstrating that reward-driven temperature control not only improves the most likely solution but also broadens the effective search space for discovering multiple correct reasoning paths.

\textbf{Out-of-Domain Generalization.}
To assess whether the learned temperature policy captures domain-general reasoning patterns, we evaluate on three Out-of-Domain (OOD) tasks: GPQA-Diamond, MMLU-Pro, and HumanEval in \cref{tab:extended_generalization_percent}. IntroLLM demonstrates consistent improvements across all OOD tasks. On the 1.7B model, IntroLLM surpasses the second-best baseline by 1.26\% on GPQA, 0.23\% on MMLU-Pro, and 1.82\% on HumanEval. The 4B model shows similar patterns, with gains of 0.07\% on GPQA, 0.49\% on MMLU-Pro, and 0.61\% on HumanEval. The consistent gains across diverse domains, ranging from scientific reasoning to code generation, demonstrate that IntroLLM's token-level temperature adaptation captures generalizable exploration strategies.

\begin{table}[t]
    \centering
    \caption{\textbf{Out-of-domain generalization performance.} We report accuracy (\%) for GPQA-Diamond (GPQA-D) and MMLU-Pro, and Pass@1 (\%) for HumanEval.}
    \label{tab:extended_generalization_percent}
    \small
    \setlength{\tabcolsep}{3pt}
    \begin{tabular}{lccc}
        \toprule
        \textbf{Model / Method} & \textbf{GPQA-D} & \textbf{MMLU-Pro} & \textbf{HumanEval} \\
        \midrule
        \multicolumn{4}{c}{\textit{\textbf{Qwen3-1.7B-Base}}} \\
        \midrule
        Base & 21.72 & 15.49 & 28.05 \\
        GRPO ($\tau=1.0$) & 28.48 & 28.39 & 33.54 \\
        EAD (1.2$\rightarrow$0.1) & 29.36 & 28.66 & 33.54 \\
        Tampo (0.6-1.5) & 27.08 & 27.16 & 35.98 \\
        \textbf{IntroLLM (Ours)} & \textbf{30.62} & \textbf{28.89} & \textbf{37.80} \\
        \midrule
        \multicolumn{4}{c}{\textit{\textbf{Qwen3-4B-Base}}} \\
        \midrule
        Base & 26.77 & 25.06 & 66.46 \\
        GRPO ($\tau=1.0$) & 35.20 & 36.60 & 75.61 \\
        EAD (1.2$\rightarrow$0.1) & 38.19 & 36.22 & 76.22 \\
        Tampo (0.6-1.5) & 36.05 & 37.23 & 76.22 \\
        \textbf{IntroLLM (Ours)} & \textbf{38.26} & \textbf{37.72} & \textbf{76.83} \\
        \bottomrule
    \end{tabular}
    \vspace{-10pt}
\end{table}

\textbf{Computational Efficiency.}\quad
\cref{tab:efficiency_throughput_1.7b} demonstrates that IntroLLM introduces negligible computational overhead. The temperature policy head adds only 21M parameters (0.122\% increase) to Qwen3-1.7B-Base, resulting in minimal memory footprint during training and inference. We measure FLOPs and generation throughput across context lengths from 1k to 24k tokens. FLOPs remain virtually identical to the baseline across all contexts. Throughput degradation is minimal: approximately 2--3 tokens/s at the common 4k context length, with negligible impact even at 24k tokens. The training time also remains efficient—the 1.7B model training completes in approximately 3h 29min with \textit{IntroLLM}, compared to 3h 5min for the standard GRPO baseline. These results demonstrate that IntroLLM achieves substantial reasoning improvements with minimal computational cost, making it practical for large-scale deployment.

\begin{table*}[t]
    \centering
    \caption{\textbf{Efficiency profile on Qwen3-1.7B-Base.} We compare the computational cost (FLOPs) and generation throughput (tokens/s) across context lengths from 1k to 24k. The \textbf{Param. $\Delta$} column indicates the additional parameter overhead introduced by IntroLLM.}
    \label{tab:efficiency_throughput_1.7b}
    \vspace{-5pt}
    
    \small
    \setlength{\tabcolsep}{7pt} 
    \renewcommand{\arraystretch}{1.1} 
    
    \begin{tabular}{llc cccccc}
        \toprule
        \textbf{Metrics} & \textbf{Method} & \textbf{Param. $\Delta$} & \textbf{1k} & \textbf{2k} & \textbf{4k} & \textbf{8k} & \textbf{16k} & \textbf{24k} \\
        \midrule
        
        \multirow{2}{*}{FLOPs} 
        & Default Sampling & -- & 4.85e+15 & 8.08e+15 & 1.45e+16 & 2.75e+16 & 5.33e+16 & 7.92e+16 \\
        & IntroLLM (Ours) & +0.122\% & 4.85e+15 & 8.09e+15 & 1.46e+16 & 2.75e+16 & 5.34e+16 & 7.93e+16 \\
        \midrule
        
        \multirow{2}{*}{\shortstack[l]{Throughput}} 
        & Default Sampling & -- & 146.37 & 145.39 & 144.14 & 143.44 & 142.14 & 139.87 \\
        & IntroLLM (Ours) & +0.122\% & 144.08 & 142.58 & 141.71 & 140.22 & 136.01 & 133.37 \\
        
        \bottomrule
    \end{tabular}
    \vspace{-10pt}
\end{table*}

\subsection{Further Analysis}

\begin{figure}[t]
    \centering
    \includegraphics[width=1.0\linewidth]{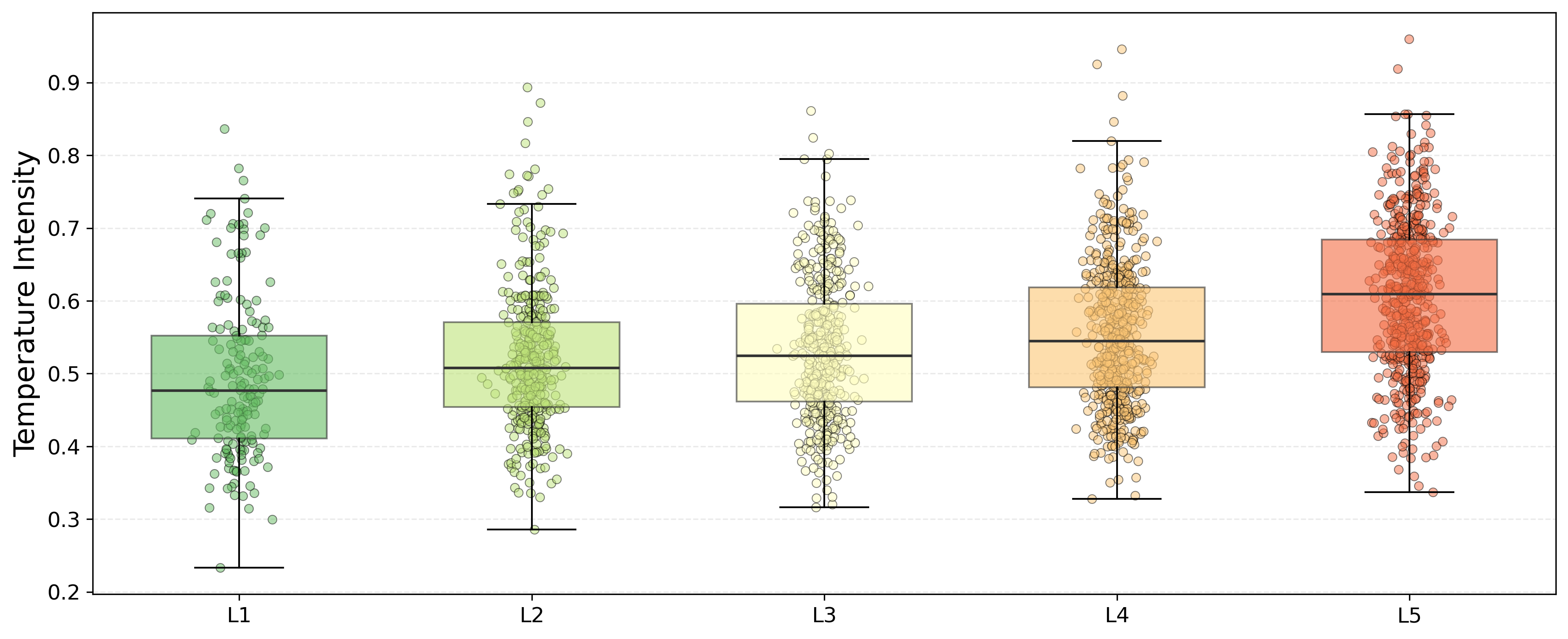}
    \caption{\textbf{Distribution of predicted temperatures across MATH-500 difficulty levels (L1–L5, from easy to hard).} As the problems get harder, the average temperature tends to increase.}
    \label{fig:MATH500_difficulty_temperature}
    \vspace{-10pt}
\end{figure}

\textbf{Difficulty-Aware Temperature Scaling.}
To investigate whether the learned policy adapts to task complexity, we analyze predicted temperatures across MATH-500 difficulty levels (L1--L5). As shown in \cref{fig:MATH500_difficulty_temperature}, median temperature increases monotonically from L1 to L5, demonstrating a clear positive correlation between problem difficulty and exploration intensity. This indicates the policy autonomously recognizes task difficulty from hidden states and allocates larger exploration budgets to harder problems—a key driver of performance gains on challenging benchmarks like AIME and Omni-Math.

\begin{figure}
    \centering
    \includegraphics[width=1.0\linewidth]{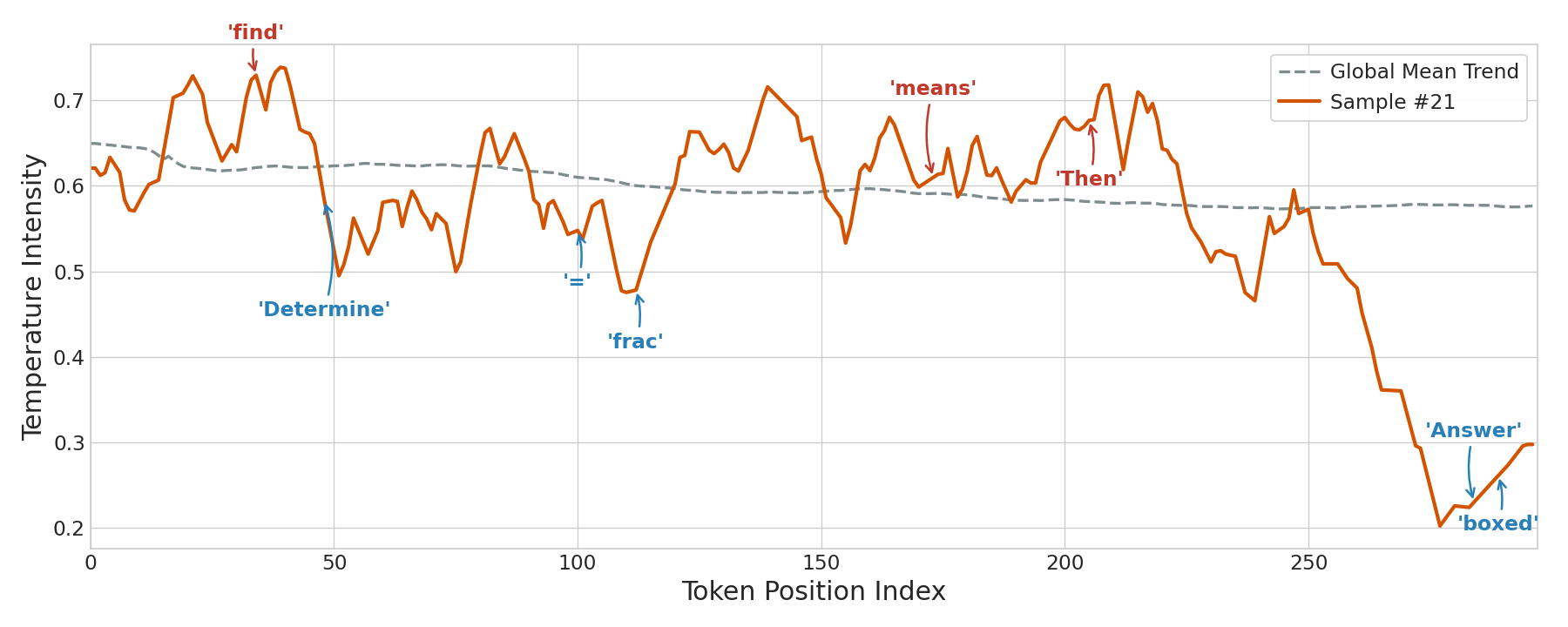}
    \caption{\textbf{Learned temperature patterns.} Gray dashed line: global average across MATH-500 showing natural annealing. Orange line: individual problem showing "reasoning rhythm" with peaks at logical pivots and valleys during computation.}
    \label{fig:cot_temp_change}
    \vspace{-10pt}
\end{figure}

\textbf{Temperature Dynamics During Reasoning.}
To understand IntroLLM's inference-time behavior, we visualize per-token temperatures on MATH-500 (\cref{fig:cot_temp_change}). Globally (averaged across all problems), the model exhibits a natural annealing pattern, shifting from early exploration to late exploitation. At the individual problem level, IntroLLM develops a "reasoning rhythm"—modulating temperature through multiple peaks and troughs, with higher temperatures at logical pivots and lower temperatures during computation and answer synthesis.

\begin{table*}[t]
    \centering
    \caption{\textbf{Ablation study on different temperature strategies.} We report \textbf{Avg@8 (A@8)} and \textbf{Pass@8 (P@8)} performance (\%).}
    \label{tab:ablation}
    \small
    \setlength{\tabcolsep}{3pt}
    \begin{tabular}{l cccccccc p{0.1cm} cccccccc}
        \toprule
        \multirow{3}{*}{\textbf{Method}} & \multicolumn{8}{c}{\textbf{Qwen-1.7B-Base}} && \multicolumn{8}{c}{\textbf{Qwen-4B-Base}} \\
        \cmidrule(lr){2-9} \cmidrule(lr){11-18}
        & \multicolumn{2}{c}{\textbf{AIME 24}} & \multicolumn{2}{c}{\textbf{AMC23}} & \multicolumn{2}{c}{\textbf{MATH}} & \multicolumn{2}{c}{\textbf{Average}} && \multicolumn{2}{c}{\textbf{AIME 24}} & \multicolumn{2}{c}{\textbf{AMC23}} & \multicolumn{2}{c}{\textbf{MATH}} & \multicolumn{2}{c}{\textbf{Average}} \\
        \cmidrule(lr){2-3} \cmidrule(lr){4-5} \cmidrule(lr){6-7} \cmidrule(lr){8-9} \cmidrule(lr){11-12} \cmidrule(lr){13-14} \cmidrule(lr){15-16} \cmidrule(lr){17-18}
        & \textbf{A@8} & \textbf{P@8} & \textbf{A@8} & \textbf{P@8} & \textbf{A@8} & \textbf{P@8} & \textbf{A@8} & \textbf{P@8} && \textbf{A@8} & \textbf{P@8} & \textbf{A@8} & \textbf{P@8} & \textbf{A@8} & \textbf{P@8} & \textbf{A@8} & \textbf{P@8} \\
        \midrule
        Prompt-level & 7.08 & 20.00 & 35.31 & 67.50 & 65.35 & 85.40 & 35.91 & 57.63 && 17.92 & 33.33 & 49.38 & 82.50 & \textbf{80.97} & 92.40 & 49.42 & 69.41 \\
        Always-update & 7.08 & 23.33 & \textbf{39.69} & 67.50 & 66.75 & 84.80 & 37.84 & 58.54 && 16.25 & 36.67 & 47.19 & 85.00 & 80.17 & 91.80 & 47.87 & 71.16 \\
        \midrule
        \textbf{IntroLLM} & \textbf{9.58} & \textbf{26.67} & 39.06 & \textbf{72.50} & \textbf{67.53} & \textbf{86.20} & \textbf{38.72} & \textbf{61.79} && \textbf{19.17} & \textbf{40.00} & \textbf{52.50} & \textbf{87.50} & 80.73 & \textbf{93.00} & \textbf{50.80} & \textbf{73.50} \\
        \bottomrule
    \end{tabular}
    \vspace{-10pt}
\end{table*}

\begin{figure}[t]
    \centering
    \includegraphics[width=0.8\linewidth]{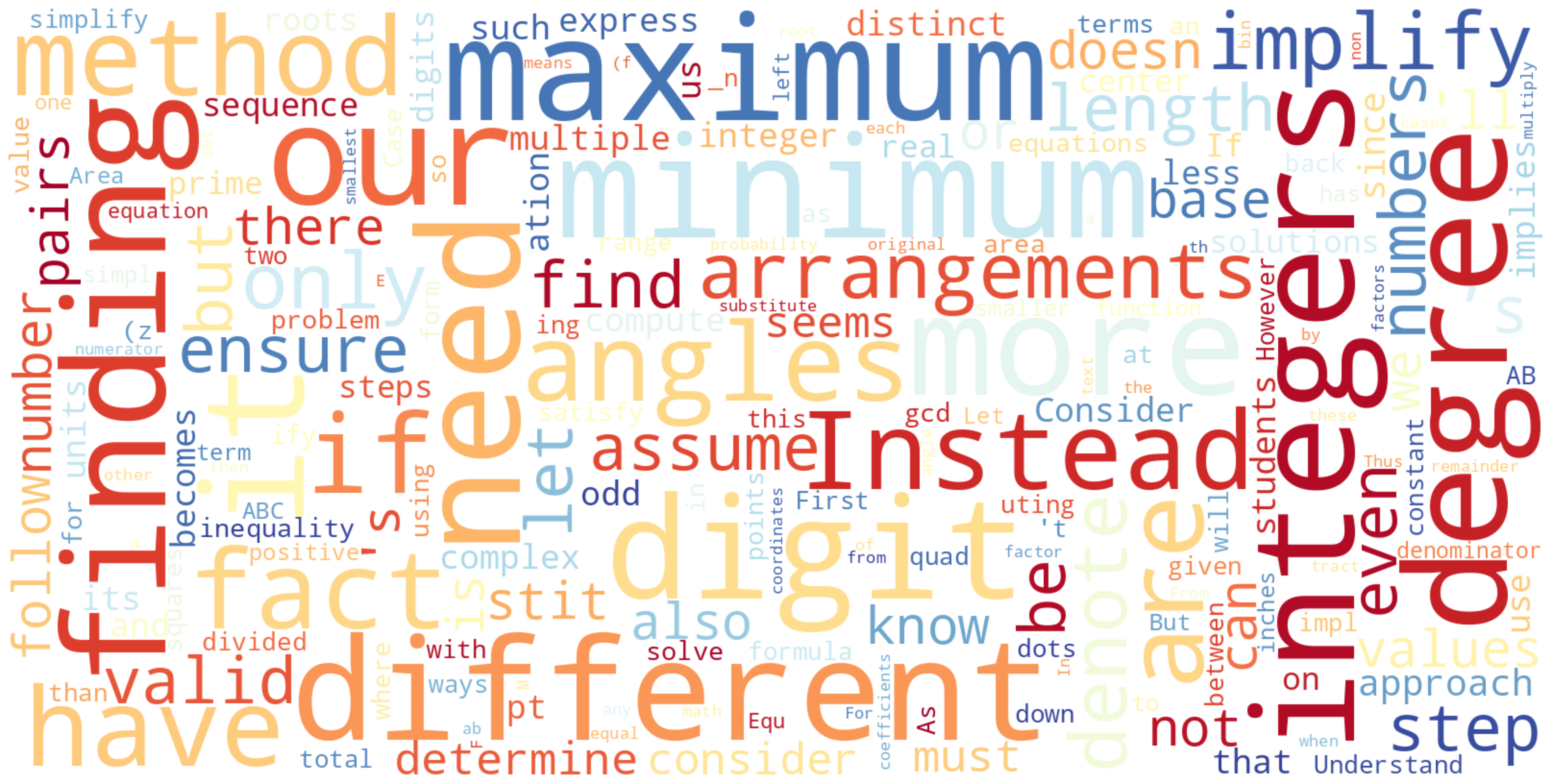}
    \caption{\textbf{Reasoning keywords trigger high temperatures.} Wordcloud of top 100 highest-temperature tokens across MATH-500. Reasoning keywords like "assume", "consider", and "finding" consistently receive increased exploration.}
    \label{fig:wordcloud}
    \vspace{-10pt}
\end{figure}

\textbf{Targeted Exploration at Reasoning Keywords.}
To identify semantic patterns behind temperature spikes, we analyze top 100 highest-temperature tokens across all MATH-500 trajectories. \cref{fig:wordcloud} shows that IntroLLM concentrates higher temperatures on reasoning-critical keywords such as "assume", "consider", "finding", and "different". This learned behavior aligns with recent empirical findings that a small set of critical tokens drives exploration in reasoning chains~\cite{wang2025beyond_80_20}, demonstrating that IntroLLM automatically discovers effective exploration strategies through reward-driven learning.

\begin{figure}
    \centering
    \includegraphics[width=1.0\linewidth]{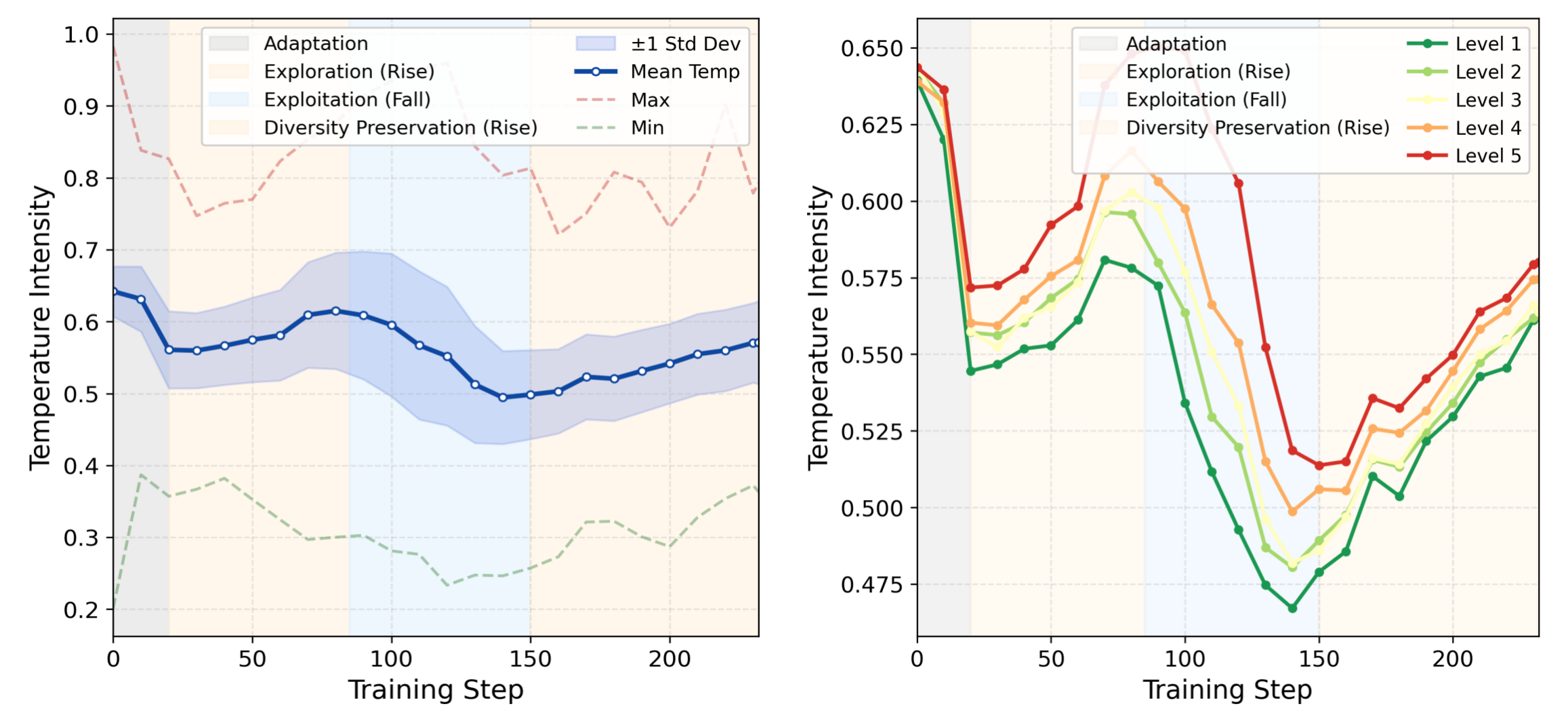}
    \caption{\textbf{Temperature intensity evolution during training.} (Left) Global mean and extrema trajectories. (Right) Mean values across difficulty levels. The policy follows an emergent, non-monotonic cycle of exploration, exploitation, and diversity preservation, distinct from traditional annealing schedules.}
    \label{fig:train_temp_evolution}
    \vspace{-10pt}
\end{figure}

\textbf{Temperature Evolution During Training.}
To examine how IntroLLM modulates exploration over training, we visualize global temperature evolution in \cref{fig:train_temp_evolution}. The model exhibits a non-monotonic adaptation pattern contrasting sharply with traditional annealing schedules. The policy initially increases temperature to enable broad exploration, then cools to consolidate high-reward patterns, and finally rises again in late training to preserve solution diversity. This emergent "exploration-exploitation-diversity" cycle, absent in pre-defined schedules, explains the superior Pass@8 performance and demonstrates that the policy learns to balance accuracy and diversity throughout training.

\begin{figure}
    \centering
    \includegraphics[width=1\linewidth]{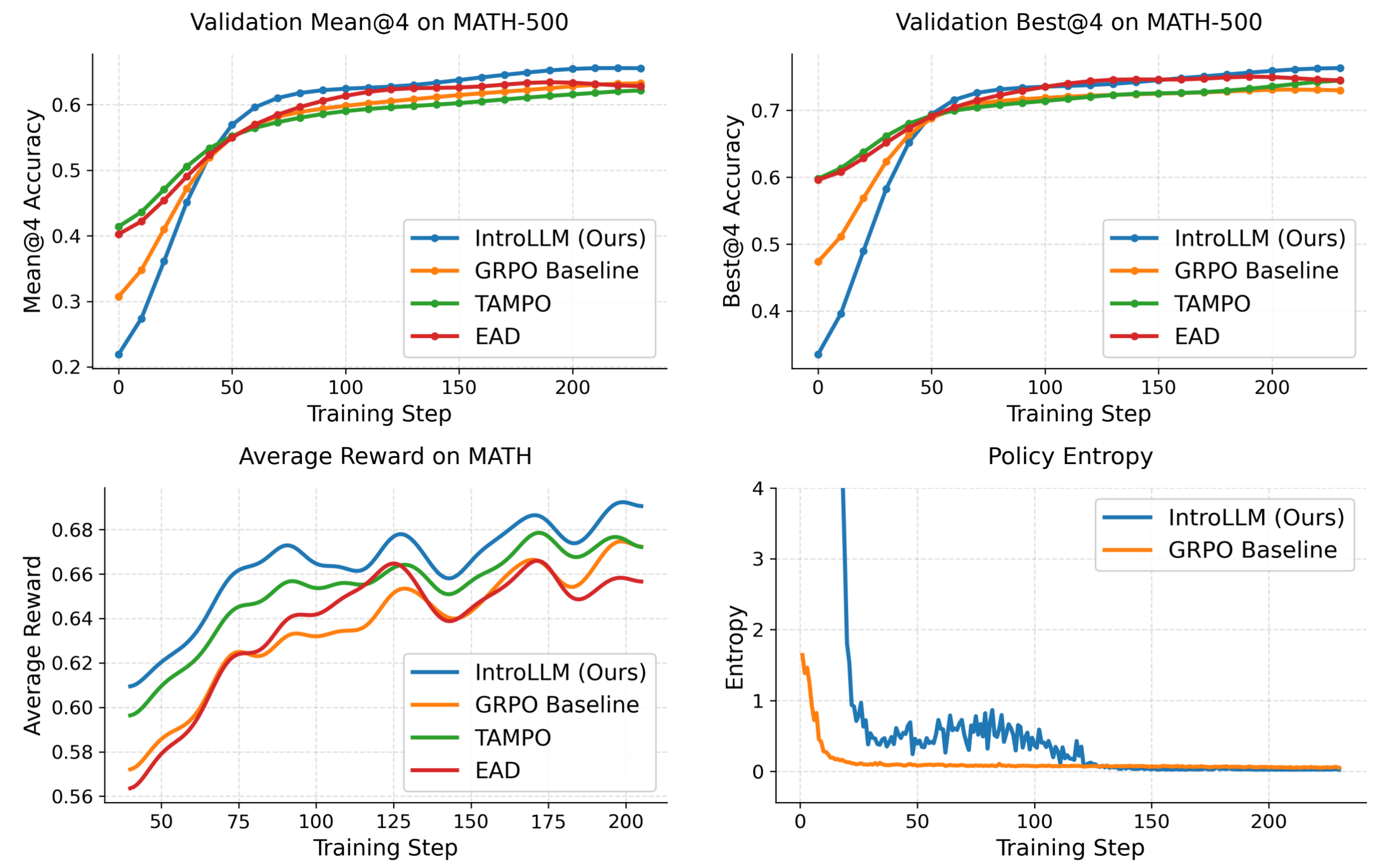}
    \caption{\textbf{Training curves on Qwen3-1.7B-Base.} IntroLLM maintains significantly higher policy entropy than the GRPO. This sustained exploration leads to higher average rewards and superior validation performance in both Mean@4 and Best@4 metrics.}
    \label{fig:training_curves}
    \vspace{-10pt}
\end{figure}

\textbf{Mitigating Entropy Collapse.}\quad
We monitor policy entropy and rewards to assess training stability (\cref{fig:training_curves}). Standard GRPO suffers from rapid entropy collapse, leading to premature convergence. In contrast, IntroLLM maintains significantly higher entropy during early training, facilitating more thorough exploration of the reasoning space. This sustained exploration translates to consistently higher average rewards and superior validation performance (Mean@4 and Best@4). These results demonstrate that adaptive temperature control effectively mitigates entropy collapse and stabilizes RLVR training.

\subsection{Ablation Study}


We ablate two key design choices of IntroLLM's temperature policy: (i) control granularity (prompt-level vs. token-level), and (ii) update frequency (selective vs. every-token). \cref{tab:ablation} compares three variants:

\textbf{Prompt-level} predicts a single temperature for the entire response based on the prompt, achieving 35.91\%/49.42\% average Avg@8 (1.7B/4B). This coarse-grained control cannot adapt to reasoning dynamics within a sequence.

\textbf{Always-update} samples a new temperature at every token without the Bernoulli trigger, achieving 37.84\%/47.87\%. While offering maximum flexibility, this introduces excessive variance that destabilizes learning—on 4B, it underperforms even prompt-level control by 1.55\%, despite being more fine-grained.

\textbf{IntroLLM} uses selective token-level updates via the mixed discrete-continuous action space, achieving 38.72\%/50.80\%. By learning when to update temperature (Bernoulli) and how much (Beta), IntroLLM outperforms prompt-level by 2.81\%/1.38\% and always-update by 0.88\%/2.93\%.

These results demonstrate that both token-level granularity and selective updates are essential—token-level control enables adaptation to reasoning dynamics, while selective updates avoid unnecessary variance.

\section{Related Work}

\textbf{RLVR and Exploration Control.}
LLM post-training has evolved from outcome-level preference modeling~\cite{christiano2017deep,ouyang2022training,rafailov2023direct} to process-level reasoning rewards~\cite{lightman2023let,cui2025process,choudhury2025process}. To mitigate reward hacking in complex tasks~\cite{gao2023scaling,wang2025beyond_reward_hacking}, Reinforcement Learning from Verifiable Rewards (RLVR)~\cite{jaech2024openai,guo2025deepseek} has emerged as the standard for reasoning-oriented models, driving research into scalable optimization and reward design~\cite{yu2025dapo,guo2025segment,zheng2025group,chen2025minimax,yuan2024free,zhao2025absolute}. However, the sparse feedback in RLVR makes discovering valid reasoning paths particularly challenging. While traditional exploration relies on entropy maximization~\cite{ziebart2008maximum,haarnoja2018soft} or intrinsic motivation~\cite{burda2018exploration,pathak2017curiosity}, recent RLVR studies address this by localizing exploration to decision-critical tokens~\cite{cheng2025reasoning,cui2025entropy,wang2025beyond_80_20,he2025skywork,wang2025reinforcement}, optimizing Pass@k objectives~\cite{chen2025pass,walder2025pass}, or incorporating semantic diversity signals~\cite{hu2025diversity,li2025jointly}. Unlike these heuristic-based approaches, IntroLLM treats exploration control itself as a learnable, reward-driven policy.

\textbf{Adaptive Temperature Sampling.}
While temperature effectively modulates policy entropy, static schedules fail to account for token-level heterogeneity and prompt-specific dynamics. Empirical studies confirm that optimal exploration requirements vary significantly across training stages and token positions~\cite{an2025polaris,yang2025let}. Recent dynamic strategies either incorporate parametric control via supervised fine-tuning~\cite{wang2025end} or adjust distribution through heuristics such as entropy or KL-divergence~\cite{zhang2024edt,zhuang2025exploring,chang2023kl}. Unlike these heuristics, IntroLLM optimizes temperature as a learnable policy from task-level rewards, enabling a fully automated, reward-driven exploration.

\section{Conclusion}
In this work, we studied the role of sampling temperature in reinforcement learning with verifiable rewards and argued that exploration control should be treated as a first-class learning problem rather than a fixed heuristic. We introduced a hierarchical formulation in which sampling temperature is modeled as a learnable policy and optimized jointly with token generation using the same task reward. This framework enables context-aware, reward-driven exploration that adapts across prompts, token positions, and training stages. Empirical results demonstrate that learning temperature leads to more effective exploration, improved reasoning performance, and more stable training compared to fixed or heuristically scheduled temperatures. Beyond the specific instantiation studied here, our formulation suggests a broader perspective on decoding as a controllable and learnable component of RL-based language model training, opening avenues for principled optimization of other generation-time decisions.




\section*{Impact Statement}
This work contributes to the study of reinforcement learning–based post-training for large language models by proposing a principled framework for adaptive exploration control. By learning when and how much to explore during generation, our method can improve reasoning reliability and training efficiency in settings with verifiable rewards. This has the potential to benefit applications that rely on structured reasoning and correctness, such as mathematics, scientific problem solving, code generation, and formal verification, where discovering valid solution trajectories under sparse feedback is critical.

From a methodological perspective, treating decoding parameters as learnable control variables may encourage more transparent and systematic design of RL-based language model training, reducing reliance on brittle heuristics and extensive manual tuning. Improved exploration efficiency could also lower the computational cost of RLVR training, potentially reducing energy consumption and resource barriers for research and deployment.

At the same time, our work does not directly address broader risks associated with large language models, such as misuse, bias, or hallucination outside verifiable settings. While adaptive exploration can improve reasoning diversity, it may also increase the generation of incorrect or unsafe intermediate outputs if applied without reliable verification signals. As a result, the proposed framework is most appropriate in domains where correctness can be automatically checked, and caution is required when extending it to open-ended or safety-critical applications.

Overall, this work advances the technical understanding of exploration in RL-based language model training, with potential positive impact on reasoning robustness and efficiency, while highlighting the continued importance of verification, evaluation, and responsible deployment.

\bibliography{icml2026}
\bibliographystyle{icml2026}

\newpage
\appendix
\onecolumn

\section{Training Algorithm}
\label{app:algorithm}

\begin{algorithm}[h]
   \caption{Introspective LLM Training via Hierarchical RL}
   \label{alg:introllm_final}
\begin{algorithmic}[1]
   \STATE {\bfseries Input:} Token policy $\pi_\theta$, temperature policy $\pi_\phi$, dataset $\mathcal{D}$, group size $G$, clipping parameter $\epsilon$, bounds $[\tau_{\min}, \tau_{\max}]$.
   \STATE {\bfseries Initialize:} $\theta \leftarrow \theta_{\text{pre}}$, $\phi \leftarrow \phi_{\text{init}}$, $\theta_{\text{old}} \leftarrow \theta, \phi_{\text{old}} \leftarrow \phi$.
   
   \FOR{each training iteration}
       \STATE Sample a batch of prompts $\{x\} \sim \mathcal{D}$.
       \FOR{each prompt $x$}
           \STATE {\color{gray} // \textit{Hierarchical Rollout}}
           \FOR{$i = 1$ {\bfseries to} $G$}
               \STATE Initialize $\tau_{i,0} = 1.0$, trajectory $\mathcal{T}_i = \emptyset$.
               \FOR{$t = 1$ {\bfseries to} $T$}
                   \STATE Compute hidden state $h_{i,t}$ from prefix $[x, y_{i,<t}]$.
                   \STATE {\color{gray} // \textit{Step 1: Temperature Selection (Policy $\pi_\phi$)}}
                   \STATE Sample $c_{i,t} \sim \text{Bernoulli}(\sigma(u_{c,t}))$, where $[u_c, u_\alpha, u_\beta] = f_\phi(h_{i,t})$.
                   \IF{$c_{i,t} = 1$}
                       \STATE Sample $z_{i,t} \sim \text{Beta}(\text{softplus}(u_\alpha), \text{softplus}(u_\beta))$.
                       \STATE $\tau_{i,t} = \tau_{\min} + z_{i,t}(\tau_{\max} - \tau_{\min})$.
                   \ELSE
                       \STATE $\tau_{i,t} = \tau_{i,t-1}$.
                   \ENDIF
                   \STATE {\color{gray} // \textit{Step 2: Token Selection (Policy $\pi_\theta$)}}
                   \STATE Sample $y_{i,t} \sim \pi_\theta(\cdot \mid h_{i,t}, \tau_{i,t})$.
                   \STATE $\mathcal{T}_i \leftarrow \mathcal{T}_i \cup \{(c_{i,t}, z_{i,t}, y_{i,t}, \tau_{i,t})\}$.
               \ENDFOR
               \STATE $R_i = \text{VerifiableReward}(x, y^{(i)})$.
           \ENDFOR
           
           \STATE {\color{gray} // \textit{Group Relative Advantage Estimation}}
           \STATE $A_i = \frac{R_i - \text{mean}(\{R_k\})}{\text{std}(\{R_k\}) + 10^{-8}}$ for $i=1 \dots G$.
           
           \STATE {\color{gray} // \textit{Step A: Update Token Policy $\theta$ (Fix $\tau$)}}
           \STATE $\mathcal{L}(\theta) = \frac{1}{G} \sum_{i,t} \left[ \min(r_{i,t}^\theta A_i, \text{clip}(r_{i,t}^\theta, 1-\epsilon, 1+\epsilon) A_i) \right]$
           
           \STATE {\color{gray} // \textit{Step B: Update Temperature Policy $\phi$ (Fix $y$)}}
           \STATE $\log \pi_\phi(c_{i,t}, z_{i,t} \mid h_{i,t}) = \log P(c_{i,t}) + c_{i,t} \log p(z_{i,t} \mid \alpha, \beta)$.
           \STATE $\mathcal{L}(\phi) = \frac{1}{G} \sum_{i,t} \left[ \min(r_{i,t}^\phi A_i, \text{clip}(r_{i,t}^\phi, 1-\epsilon, 1+\epsilon) A_i) \right]$.
       \ENDFOR
       \STATE $\theta_{\text{old}} \leftarrow \theta, \phi_{\text{old}} \leftarrow \phi$.
   \ENDFOR
\end{algorithmic}
\end{algorithm}

\section{Implementation Details}
\label{app:implementation}

\subsection{Model Architecture}
The model extends the base architecture with a lightweight MLP head branching from the final hidden state to predict temperature parameters. This lightweight design allows the temperature head to serve as a plug-and-play module, making it easily adaptable to models of the same family that have undergone fine-tuning~\cite{qing2024alphalora,wang2025loralib,yin2025evaluating,feng2024continual,han2025cat} or compression~\cite{zhou2025dropping,lu2024reassessing, li2025catp}. Additionally, this design facilitates straightforward migration to other architectures, such as diffusion language models~\cite{han2025c}.

\textbf{Temperature Policy Head:}
\begin{itemize}
    \item \textbf{Structure}: Linear($d_{\text{model}}$, $d_{\text{model}}/2$) $\to$ ReLU $\to$ Linear($d_{\text{model}}/2$, 3).
    \item \textbf{Output}: 3 logits $[u_c, u_\alpha, u_\beta]$ determining the temperature update probability and Beta distribution parameters.
    \item \textbf{Initialization}: Weights are initialized with $\mathcal{N}(0, 0.02)$. To facilitate initial exploration, final layer biases are set to $u_c=0$ (implies $P(\text{change})=0.5$) and $u_\alpha=u_\beta \approx 0.541$ (implies $\text{Beta}(1,1)$, i.e., Uniform distribution). This maximum-entropy initialization allows the model to explore the full temperature range early in training.
\end{itemize}

\subsection{Hyperparameters}
We list the detailed hyperparameters for the \textit{IntroLLM} training in \cref{tab:hyperparams}. To stabilize the hierarchical optimization, we employ decoupled learning rates, setting the token policy to $1 \times 10^{-6}$ and the temperature policy to $5 \times 10^{-5}$. The training process was executed on a cluster of 8 NVIDIA H100 (80GB) GPUs using the AdamW optimizer, utilizing a global batch size of 128. We further set the group size to $G=8$ to provide reliable relative advantage estimation during the rollout phase, while the maximum sequence length was capped at 3072 tokens to balance reasoning depth and compute efficiency.

\begin{table}[h]
    \centering
    \caption{Hyperparameters for IntroLLM Training}
    \label{tab:hyperparams}
    \small
    \begin{tabular}{lc}
        \toprule
        \textbf{Parameter} & \textbf{Value} \\
        \midrule
        Token Policy LR & $1 \times 10^{-6}$ \\
        Temperature Policy LR & $5 \times 10^{-5}$ \\
        Batch Size & 128 \\
        Group Size ($G$) & 8 \\
        Temperature Bounds $[\tau_{\min}, \tau_{\max}]$ & $[0.6, 1.5]$ \\
        Max Sequence Length & 3072 \\
        Optimizer & AdamW \\
        \bottomrule
    \end{tabular}
\end{table}

\subsection{Reward Function}
We strictly follow the RLVR setting. The reward is binary: $R(x, y) = 1$ if the final answer matches the ground truth after symbolic verification, and $R(x, y) = 0$ otherwise. No partial rewards or process supervision rewards were used.

\section{Datasets}
\label{app:datasets}

\subsection{Training Set: MATH}
We utilize the training split of the \textbf{MATH} dataset \cite{hendrycks2021measuring}, which contains 7,500 mathematics problems across seven categories (e.g., Algebra, Calculus, Counting \& Probability). Each problem is accompanied by a step-by-step solution. Following standard RLVR practices, we only use the problem prompts for reinforcement learning, while the ground-truth answers serve as the basis for the verifiable reward function.

\subsection{Evaluation Benchmarks}

\textbf{MATH-500} \cite{lightman2023let} is a high-quality subset of the original MATH test set, sampled to provide a balanced representation of difficulty levels (L1--L5) and subject categories. It was first introduced by OpenAI to evaluate process-based reward models (PRMs). Given its manageable size and rigorous difficulty distribution, it serves as our primary benchmark for analyzing per-token temperature fluctuations and reasoning uncertainty.

\textbf{AIME 2024} represents the most recent iteration of the American Invitational Mathematics Examination, a premier competition for high-school students. It consists of 15 highly complex problems where the answer is an integer between 000 and 999. Unlike standard datasets, AIME requires long-context logical deduction and is less susceptible to memorization, making it an ideal testbed for evaluating the exploration capabilities of IntroLLM on "out-of-distribution" hard problems.

\textbf{AMC 2023} encompasses problems from the American Mathematics Competitions (AMC 10 and 12). These 25-question, 75-minute multiple-choice examinations are designed to challenge problem-solving skills under time constraints. We use the 2023 version to ensure that the models are evaluated on recent, potentially unobserved data. This benchmark tests the model's ability to maintain high precision while navigating the breadth of secondary school mathematics.

\textbf{Minerva Math} \cite{lewkowycz2022solving} is a specialized evaluation suite derived from the Minerva model's assessment framework. It primarily focuses on STEM-focused problem solving, drawing from sources like the MATH dataset and various university-level exams. We utilize this benchmark to evaluate the model's robustness in formal scientific reasoning and its ability to handle multi-step calculations that require consistent numerical accuracy.

\textbf{OlympiadBench} \cite{he2024olympiadbench} is a comprehensive, bilingual (English and Chinese) benchmark featuring high-level competition problems from International Mathematical Olympiads (IMO) and other prestigious contests. It includes both mathematics and physics domains. We leverage its mathematical portion to test whether the learned temperature policy can adapt to the extreme reasoning depths required for Olympiad-level deduction.

\textbf{Omni-Math} \cite{gao2024omni} is a recently released, large-scale benchmark designed to address the "saturation" of previous math datasets. It contains over 4,000 problems meticulously categorized into fine-grained sub-fields and difficulty tiers. Due to its vast coverage and high average difficulty, Omni-Math provides the most statistically significant measure of our model's general mathematical proficiency across the entire landscape of modern mathematics.





\section{Baseline Methods Details}
\label{app:baselines}

To ensure the reproducibility of our comparative study, we provide the comprehensive technical configurations and hyperparameter settings for the two primary adaptive baselines. Both methods were integrated into our GRPO framework using the official logic described in their respective original works.

\subsection{Standard GRPO (Fixed Temperature)}
We evaluate three variants of the standard \textbf{GRPO} \cite{shao2024deepseekmath} using fixed sampling temperatures to establish baseline performance bounds for static exploration:
\begin{itemize}[leftmargin=*]
    \item \textbf{GRPO ($\tau=0.6$)}: Represents an \textbf{exploitative} configuration, focusing on high-probability reasoning paths to ensure output stability.
    \item \textbf{GRPO ($\tau=1.0$)}: The default setting used in most RLVR literature, providing a neutral balance between exploration and exploitation.
    \item \textbf{GRPO ($\tau=1.2$)}: Represents an \textbf{exploratory} configuration, intentionally increasing entropy to discover diverse reasoning trajectories.
\end{itemize}
In these baselines, the temperature remains constant throughout both the training and rollout phases, and across all token positions within a sequence.

\subsection{EAD: Exploratory Annealed Decoding}
\textbf{EAD} \cite{yang2025let} focuses on \textit{intra-sequence exploration}, grounded in the intuition that early tokens in a reasoning chain~\cite{shi2024prompt} dictate the semantic direction and benefit most from high-entropy exploration, while later tokens should converge to stable exploitation.

\textbf{Mechanism:} The sampling temperature $\tau_t$ at token position $t$ is governed by a dynamic simulated annealing schedule:
\[ \tau_t = \max(\tau_{\min}, \tau_{\text{start}} \cdot \gamma^t), \]
where $\tau_{\text{start}}=1.2$ and $\tau_{\min}$ is set to $0.1$ for our 1.7B/4B models to encourage sharp convergence. To avoid disrupting the generation of standard response templates (e.g., "Let's think step by step"), we implement a \textbf{cutoff threshold} where $\tau_t$ remains fixed at $\tau_{\text{start}}$ for the first 10 tokens.

\textbf{Global Adaptation:} To account for the increasing length of reasoning chains during RL training, the decay rate $\gamma$ is dynamically adjusted relative to the current training step $n$:
\[ \gamma_n = \exp \left( -\frac{c_0}{L_{\text{avg}}} \cdot \frac{n}{N_{\text{total}}} \right), \]
where $L_{\text{avg}}$ is the moving average response length and $c_0=25$ is the initial decay coefficient. This ensures that the annealing effect scales proportionally with the model's evolving verbosity. We also employ Truncated Importance Sampling (TIS) to maintain stability under high-variance off-policy updates.

\subsection{TAMPO: Temperature Adaptive Meta Policy Optimization}
\textbf{TAMPO} \cite{dang2026temperature} treats temperature selection as a \textit{meta-policy learning} problem, optimizing a discrete distribution over potential temperatures across training iterations.

\textbf{Mechanism:} TAMPO utilizes a hierarchical two-loop optimization process. In the \textbf{Inner Loop}, the model samples a temperature $\tau_{\text{selected}}$ from the current meta-policy to generate trajectories. In the \textbf{Outer Loop}, the meta-policy is updated based on \textbf{Temperature-Specific Advantages (TSAs)}. For a rollout $y$ with reward $R$, the TSA for a candidate temperature $\tau_j$ is calculated as:
\[ \text{TSA}(\tau_j) = A \cdot \frac{P(y \mid \tau_j)}{\sum_{k} P(y \mid \tau_k)}, \]
where $A$ is the group-relative advantage. This weighting reinforces temperatures that assign higher likelihood to high-reward trajectories.

\textbf{Implementation Details:}
\begin{itemize}[leftmargin=*]
    \item \textbf{Action Space:} A discrete set of 10 candidate values $\mathcal{T} = \{0.6, 0.7, \dots, 1.5\}$.
    \item \textbf{Selection Strategy:} We apply Nucleus Sampling ($p=0.9$) over the meta-policy distribution to select $\tau$ for each batch, ensuring a balance between exploiting the best temperature and maintaining meta-exploration.
    \item \textbf{Warmup \& Stability:} The first 10\% of training steps are designated as a warmup phase where $\tau$ is fixed at $1.0$. We use an Exponential Moving Average (EMA) with $\alpha=0.9$ to smooth the TSA updates and stabilize the meta-policy evolution.
\end{itemize}

\subsection{Summary of Differences and Advantages}
The fundamental distinction between IntroLLM and the aforementioned baselines lies in the granularity of control and the source of adaptation. We summarize these key differences in \cref{tab:method_comparison} and highlight our primary advantages below.

\begin{table}[h]
    \centering
    \caption{Conceptual comparison of IntroLLM against baseline temperature control methods.}
    \label{tab:method_comparison}
    \small
    \begin{tabular}{lcccc}
        \toprule
        \textbf{Feature} & \textbf{Fixed GRPO} & \textbf{EAD} & \textbf{TAMPO} & \textbf{IntroLLM (Ours)} \\
        \midrule
        Control Level & Sequence & Token & Sequence & \textbf{Token} \\
        Adaptation Basis & Static & Heuristic & Training Iteration & \textbf{Internal State ($h_t$)} \\
        Optimization & None & None & Meta-Policy & \textbf{Hierarchical RL} \\
        Learned? & No & No & Yes (Partial) & \textbf{Yes (End-to-End)} \\
        \bottomrule
    \end{tabular}
\end{table}

\paragraph{1. Introspective vs. Context-blind.} 
While Standard GRPO and TAMPO apply a single temperature to an entire response, they ignore the fact that reasoning uncertainty fluctuates within a chain. EAD introduces token-level changes but follows a rigid, hand-crafted decay schedule. In contrast, IntroLLM is \textbf{introspective}: it observes the model's internal hidden states ($h_t$) to detect when the model is entering a high-uncertainty reasoning pivot, allowing it to "look inward" to decide how much to "explore outward."

\paragraph{2. Reward-driven Emergence vs. Heuristic Design.} 
EAD's annealing schedule and TAMPO's discrete bins rely on human intuition and manual tuning. IntroLLM treats temperature selection as a first-class control problem optimized directly by downstream task rewards. This allows the "reasoning rhythm" (as shown in \cref{fig:cot_temp_change}) to emerge naturally from the data, ensuring that exploration effort is allocated where it most contributes to finding the correct solution.

\paragraph{3. Efficiency through Mixed Action Space.} 
Unlike potential token-level baselines that might change temperature at every step (introducing massive variance), our mixed \textbf{Discrete--Continuous action space} (Bernoulli trigger + Beta sampling) allows the model to learn \textit{when} to stay stable and \textit{when} to adapt. This provides a superior balance between exploration flexibility and training stability, leading to the smoother and more effective convergence observed in our training curves (\cref{fig:training_curves}).

\section{Supplementary Experiments}
\label{app:additional_experiments}

\subsection{Analysis of Generation Length}
\label{app:gen_length}

\begin{figure}[h]
    \centering
    \includegraphics[width=0.9\linewidth]{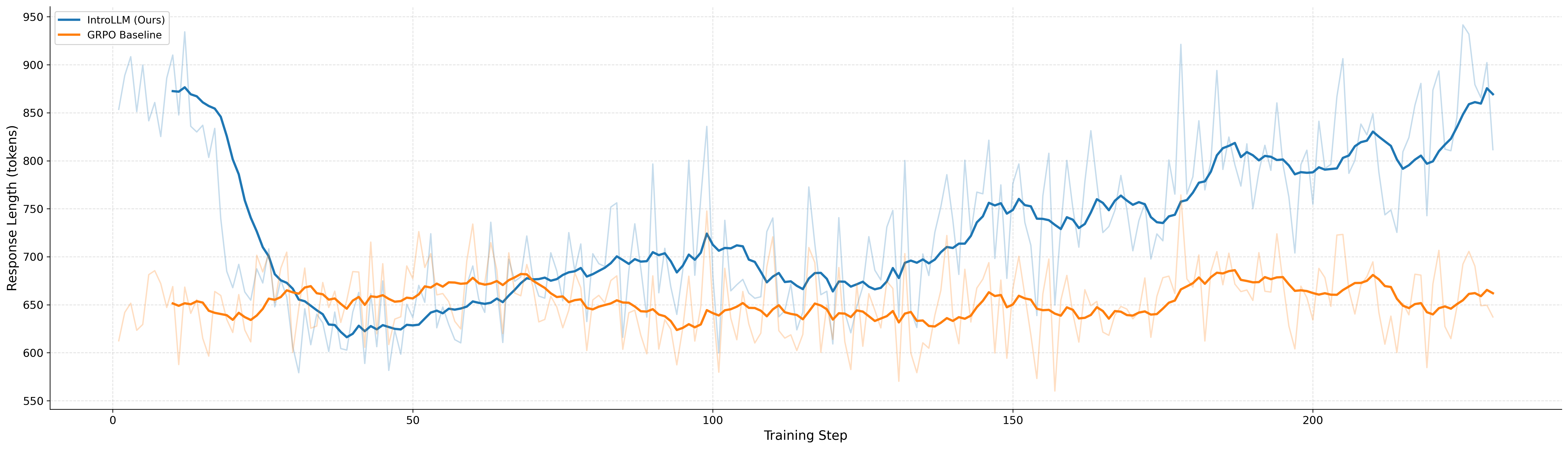}
    \caption{\textbf{Evolution of average response length during RL training.} While the GRPO baseline maintains a stagnant length of approximately 650 tokens, IntroLLM exhibits a dynamic adaptation: after an initial convergence phase, it steadily learns to generate longer, more elaborate reasoning chains to maximize task rewards.}
    \label{fig:app_cot_length}
\end{figure}

We analyzed whether learning temperature affects the verbosity or the structural complexity of the generated solutions. \cref{fig:app_cot_length} illustrates the evolution of average response length (in tokens) during the reinforcement learning process for both IntroLLM and the fixed-temperature GRPO baseline.

As shown in the figure, the standard GRPO baseline remains relatively stagnant, fluctuating around a fixed length throughout the training steps. In contrast, IntroLLM exhibits a distinct "U-shaped" adaptation followed by a consistent upward trend. Specifically, after the initial stages of training, the response length of IntroLLM steadily increases, eventually surpassing the baseline by approximately \textbf{30\%--40\%}.

We posit that this increase represents \textbf{effective thinking depth} rather than redundant verbosity. By dynamically increasing sampling temperature at critical reasoning pivots (as discussed in \cref{fig:cot_temp_change}), the model is encouraged to avoid "jumping to conclusions." Instead, it allocates more tokens to elaborate on intermediate steps and explore alternative logical paths. This emergent behavior is particularly evident in the high-difficulty benchmarks like AIME and Omni-Math, where the model's performance gain is strongly correlated with its ability to generate more detailed, multi-step verification chains. These results suggest that reward-driven temperature control allows the model to discover that \textbf{deeper elaboration under appropriate stochasticity is essential for solving complex reasoning tasks.}

\section{Declaration of LLM Usage}
\label{app:llm_usage}
We declare that Large Language Models were utilized exclusively for grammatical refinement and spell-checking to enhance the clarity and readability of this manuscript. No part of the intellectual content, experimental design, or results was generated by AI. 


\end{document}